\pdfoutput=1

\documentclass[11pt]{article}

\usepackage[preprint]{acl}
\usepackage{hyperref}
\usepackage{enumitem}
\setlist{nosep}
\usepackage{times}
\usepackage{latexsym}
\usepackage{amsmath}
\usepackage{tcolorbox}
\usepackage{booktabs}
\usepackage{multirow}
\usepackage{multicol}
\usepackage{natbib}
\usepackage[T1]{fontenc}
\usepackage[utf8]{inputenc}
\usepackage{microtype}
\usepackage{inconsolata}
\usepackage{graphicx}
\usepackage{indentfirst}

\usepackage{pifont}
\newcommand{\cmark}{\ding{51}}
\newcommand{\xmark}{\ding{55}}

\title{EconLogicQA: A Question-Answering Benchmark for Evaluating \\ Large Language Models in Economic Sequential Reasoning}

\author{
    Yinzhu Quan\thanks{These authors contributed equally to this work.} \\
    Georgia Institute of Technology \\
    Atlanta, GA 30332, USA \\
    \texttt{yquan9@gatech.edu} \\\And
    Zefang Liu$^*$ \\
    Georgia Institute of Technology \\
    Atlanta, GA 30332, USA \\
    \texttt{liuzefang@gatech.edu} \\
}

\begin{document}
\maketitle

\begin{abstract}

In this paper, we introduce EconLogicQA, a rigorous benchmark designed to assess the sequential reasoning capabilities of large language models (LLMs) within the intricate realms of economics, business, and supply chain management. Diverging from traditional benchmarks that predict subsequent events individually, EconLogicQA poses a more challenging task: it requires models to discern and sequence multiple interconnected events, capturing the complexity of economic logics. EconLogicQA comprises an array of multi-event scenarios derived from economic articles, which necessitate an insightful understanding of both temporal and logical event relationships. Through comprehensive evaluations, we exhibit that EconLogicQA effectively gauges a LLM's proficiency in navigating the sequential complexities inherent in economic contexts. We provide a detailed description of EconLogicQA dataset and shows the outcomes from evaluating the benchmark across various leading-edge LLMs, thereby offering a thorough perspective on their sequential reasoning potential in economic contexts. Our benchmark dataset is available at \url{https://huggingface.co/datasets/yinzhu-quan/econ_logic_qa}.

\end{abstract}

\section{Introduction}

Logical reasoning is a pivotal skill in many professional and academic domains, enabling individuals to make informed decisions by understanding relationships between sequential events or pieces of information. In practice, the reasoning capabilities of large language models (LLMs) are frequently utilized across various applications, yet their effectiveness in logical reasoning remains underexplored. Despite its importance, there is an evident gap in the literature regarding the capability of LLMs to perform logical reasoning at a high level. This paper addresses this gap by introducing EconLogicQA, a new benchmark designed to rigorously assess the logical reasoning capabilities of LLMs specifically within the contexts of economics, business, and supply chain management.

EconLogicQA distinguishes itself from existing benchmarks by challenging LLMs to not only identify but also logically sequence multiple interconnected events from realistic economic scenarios. This approach aims to reflect the intricate decision-making processes required in these fields, going beyond mere fact recall or simple event prediction. By focusing on the sequencing of events based on logical rather than simply chronological order, EconLogicQA inspects the LLMs' ability to engage with and understand the underlying mechanics of economic phenomena.

The benchmark utilizes a curated dataset derived from a wide range of business news articles, guiding GPT-4 to generate multi-choice questions that demand an intelligent understanding of logical connections. A rigorous human review process ensures the accuracy and appropriateness of the content, refining the dataset to enhance its practical value. Through comprehensive testing across various state-of-the-art LLMs, this paper not only demonstrates EconLogicQA's effectiveness in evaluating logical reasoning but also provides insights into the potential improvements and applications of LLMs in complex reasoning tasks.

Our contributions of this paper are as follows:
\begin{enumerate}
    \item We propose a novel benchmark, EconLogicQA, which rigorously assesses LLMs' logical reasoning capabilities within economics, business, and supply chain management.
    \item We utilize GPT-4 to generate questions and answers from business articles, ensuring high-quality, well-crafted multiple-choice questions through meticulous human review.
    \item We conduct a comprehensive evaluation of both open and proprietary LLMs to assess their performance on this benchmark.
\end{enumerate}

\section{Related Work}

\textbf{Sequential Reasoning Benchmarks.} In the realm of assessing complex reasoning abilities, \citet{jin2023cladder} introduce the CL\text{\small ADDER} dataset, exploring capacities of large language models (LLMs) for causal reasoning, differentiating itself by focusing on formal rules-based causal inference instead of the typical evaluation of commonsense causality \citep{talmor2018commonsenseqa,huang2019cosmos} in Natural Language Processing (NLP). \citet{wang2023steps} present STEPS, a rigorous benchmark designed to assess models' understanding of action sequence order in sequential tasks such as cooking and manufacturing, which highlights challenges of current LLMs in performing order reasoning without specific tuning. In examining adjacent domains, \citet{guha2024legalbench} launch L\text{\small EGAL}B\text{\small ENCH}, emerging as a notable benchmark that evaluates LLMs in legal reasoning, having been developed collaboratively with legal experts to cover various facets of practical and theoretical legal analysis. \citet{yang2024aqa} establish AQA-Bench, serving as an interactive benchmark that evaluates LLMs' sequential reasoning abilities across various algorithmic tasks, including Depth-First Search (DFS), Breadth-First Search (BFS), and binary search, by requiring models to dynamically interact with the task environment, and thereby uncovering notable performance disparities among different LLMs. \citet{valmeekam2024planbench} create PlanBench as an extensible benchmark focused on evaluating LLMs' planning and reasoning capabilities, particularly in action and change, where diverse scenarios are used from the automated planning community to discern between genuine planning abilities and mere retrieval from pre-trained knowledge.

\textbf{Economic Benchmarks.} In the finance domain, \citet{shah2022flue} launch the Financial Language Understanding Evaluation (FLUE) benchmark alongside the Financial LANGuage (FLANG) model, offering a comprehensive suite of evaluations focused on economic and financial domains, significantly outperforming existing models on various NLP tasks. \citet{hendrycks2020measuring} compile the Massive Multitask Language Understanding (MMLU) benchmark of 57 diverse tasks, including the economics subject, designed to evaluate the multitask accuracy of language models, revealing that even the largest models still struggle with expert-level performance and have inconsistent accuracy across subjects. \citet{lu2023bbt} propose the BBT-CFLEB benchmark, supporting advanced understanding and generation tasks in the financial domain and fostering significant research and development in this specialized area. \citet{zhang2023fineval} present the FinEval, a specialized benchmark for assessing financial knowledge in LLMs, demonstrating significant potential through GPT-4's high performance across diverse prompt types. \citet{van2023evaluating} introduce the EconQA, a novel dataset for assessing LLMs on multiple-choice economics questions, reveals that Chain-of-Thought reasoning improves performance, particularly in mathematical queries, while prompt variations have a moderate effect on accuracy.

\section{EconLogicQA}

In this section, we detail the dataset generation and human review processes for creating the EconLogicQA benchmark and provide illustrative examples from it.

\subsection{Dataset Generation}

To streamline the question-generation process and reduce the subjectivity, labor-intensiveness, and randomness of manual creation, we utilize the GPT-4 to automatically generate questions by extracting key points from news articles. We specifically select economics-related articles from the 2011 to 2022 news dataset available on Kaggle\footnote{\url{https://www.kaggle.com/datasets/hadasu92/cnn-articles-after-basic-cleaning/}}, which is under the CC0 Public Domain license. This cleaned dataset provides a comprehensive range of economic news articles, and we further narrow our focus to those categorized under business to align with our research scope in economics.

In the data generation process, instructional prompts are developed to guide GPT-4 in creating multi-choice questions that challenge models to logically sequence events within the framework of business-related scenarios. These questions start with a brief scenario description and involve four events that must be ordered based on their logical or chronological sequence rather than their appearances in the source articles. The selected events pertain to typical business or economic situations, necessitating a deep understanding of business practices and economic principles for accurate sequencing. The prompts specify that the generated content should be original, concise, and crafted without referencing the original news articles and unnecessary details. Each question is designed to be completed independently, making it suitable for evaluation. The formatted output includes a scenario description followed by four choices labeled A, B, C, and D, concluding with the correct sequence and a brief explanation to ensure that the reasoning behind the sequence is clear and deducible solely from the information presented in the question and choices. This structure is intended to enhance comprehension and application of business concepts. See Appendix \ref{sec:prompt} Figure \ref{fig:prompt}
for an example of GPT-4 response with the prompt.

\subsection{Review Process}

In order to maintain the integrity and quality of the dataset, human verification is incorporated into the workflow. This manual review is essential as some generated responses exhibit errors in the correct sequencing of events. Each question undergoes meticulous examination, and adjustments are made to ensure accuracy and clarity in the logical sequence provided. Furthermore, the dataset undergoes a rigorous review to identify and exclude sensitive news articles that could be inappropriate. In total, 204 questions are removed from the initial pool of 854 questions. The criteria for removal include scenarios that yield multiple valid sequences and instances where a logical sequence cannot be clearly established. This comprehensive vetting process significantly enhances the evaluation quality. The final dataset consists of \textbf{650 questions}, divided into training, validation, and test sets containing 390, 130, and 130 questions, respectively.

\subsection{Dataset Examples}

To provide a clear depiction of EconLogicQA’s contents, we present two examples from the dataset in Table \ref{tab:examples}. The first example details a sequence of decisions by Costco to manage its chicken supply chain effectively, while the second outlines steps taken by the Federal Reserve to navigate fiscal challenges. These examples illustrate the dataset's primary objective: to evaluate the capability of large language models in sequencing economic events logically, not just chronologically. Each question is meticulously designed to challenge models to demonstrate their understanding of complex economic interactions and to apply logical reasoning within real-world business contexts.

\begin{table}[!h]
    \begin{tabular}{p{.95\linewidth}}
    \toprule
       \textbf{EconLogicQA Example 1} \\
    \midrule
       \textbf{Question:} Costco, a major retailer, is taking steps to control its chicken supply chain to maintain the price of its popular rotisserie chickens. Arrange the following events in the logical sequence based on Costco's strategy. \\
       \textbf{Choices:} \\
       \textbf{A.} Costco recruits farmers to grow chickens for them. \\
       \textbf{B.} Costco opens a \$450 million poultry complex in Nebraska. \\
       \textbf{C.} Costco faces supply challenges and cost pressures in the poultry industry. \\
       \textbf{D.} Costco decides to keep its rotisserie chickens at \$4.99 despite competitors increasing their prices. \\
       \textbf{Correct Order:} D, C, A, B \\
    \midrule
       \textbf{EconLogicQA Example 2} \\
    \midrule
       \textbf{Question:} Arrange the following events related to the Federal Reserve's monetary policy in their correct order.\\
       \textbf{Choices:} \\
       \textbf{A.} The Federal Reserve announces the end of its pandemic-era stimulus and accelerates the roll-back of its asset purchases.\\
       \textbf{B.} The Federal Reserve signals it would hike interest rates multiple times throughout 2022.\\
       \textbf{C.} The Federal Reserve reduces its massive balance sheet.\\
       \textbf{D.} The Federal Reserve raises interest rates due to inflation well above 2\% and a strong labor market.\\
       \textbf{Correct Order:} A, B, D, C\\
    \bottomrule
    \end{tabular}
    \caption{Question and answer examples from EconLogicQA.}
    \label{tab:examples}
\end{table}

\section{Experiments}

This section outlines experiments with the EconLogicQA dataset, assessing the sequential reasoning of multiple open and proprietary large language models (LLMs) in economic scenarios.

\subsection{Experiment Setup}

We implement experiments on various LLMs using the EconLogicQA dataset to assess its sequential reasoning capabilities within the intricate realms of economics, business, and supply chain management. We select the current mainstream open and proprietary LLMs in our study, including Llama-2 \cite{touvron2023llama1,touvron2023llama}, Llama 3 \cite{meta_llama3}, Gemma \cite{team2024gemma}, Mistral \cite{jiang2023mistral}, Yi \cite{young2024yi}, Zephyr \cite{tunstall2023zephyr}, GPT-3.5 \cite{ouyang2022training}, and GPT-4 \cite{achiam2023gpt}. Each model is evaluated in both 1-shot and 5-shot settings. We do not include 0-shot in this section because the results are unsatisfactory due to lack of examples. Therefore, we choose to using a few-shot approach for sorting problems. Accuracy is the primary metric used, offering a direct measure of each model's understanding of the concepts within the EconLogicQA dataset.

All experiments are conducted on NVIDIA A100 GPUs. Each open LLM use in this paper is sourced from the Huggingface Transformers library \cite{wolf2019huggingface}. Language Model Evaluation Harness \cite{eval-harness} is used to test open LLMs on sequential reasoning evaluation tasks. The YAML configuration file is used to set key parameters in our scenario, such as terminating outputs, extracting answers, evaluating results, and so on. LLMs are configured to stop generating responses once it begins to pose new questions after answering the question in the prompt, setting the temperature to 0 without sampling. Then, we extract the answer from the response generated by LLMs using regular expression. Finally, we verify the answer through exact matching and use accuracy as the evaluation metric.

\subsection{Experiment Results}

The results from our experiments shown in Table \ref{tab:results} indicate diverse performances among the models, with significant disparities observed in their ability to handle the sequencing of economic events. GPT-4-Turbo exhibits the highest accuracy, achieving 56.92\% in the 1-shot scenario and 56.15\% in the 5-shot scenario, making it the best-performing model in our tests. GPT-4 follows closely, demonstrating the second-highest performance with 55.38\% in the 1-shot and 53.85\% in the 5-shot settings. Remarkably, the 1-shot scenario generally results in better performance than the 5-shot scenario for these two models, which could be attributed to the models' ability to leverage their pre-trained knowledge effectively without the potential confusion introduced by additional context in the 5-shot scenario.

Open LLMs are evaluated, showing varied performances as outlined in Table \ref{tab:results}. Notably, Llama-3-8B-Instruct and Llama-3.1-8B-Instruct demonstrates significant improvements when fine-tuned with instructions compared with Llama-3-8B and Llama-3.1-8B respectively, achieving 34.62\% and 36.15\% accuracy in the 1-shot setting and 37.69\% and 36.92\% in the 5-shot setting, which highlights the substantial impact of instruction tuning on enhancing performance for question answering tasks. Similarly, Mistral-7B-Instruct-v0.3 exhibits promising results, with the accuracy of 39.23\% in the 1-shot setting and 35.38\% in the 5-shot setting, underscoring its adaptability to complex reasoning tasks, though it still lags behind GPT-4’s overall performance.

\begin{table}[h!]
\centering
\begin{tabular}{lrr}
\toprule
\textbf{Model} & \textbf{1-Shot} & \textbf{5-Shot} \\
\midrule
Llama-2-7B & 0.77\% & 1.54\% \\
Llama-2-7B-Chat & 9.23\% & 10.00\% \\
Llama-2-13B & 9.23\% & 1.54\% \\
Llama-2-13B-Chat & 14.62\% & 8.46\% \\
Llama-3-8B & 23.85\% & 23.85\% \\
Llama-3-8B-Instruct & 34.62\% & 37.69\%\\
Llama-3.1-8B & 31.54\% & 27.69\% \\
Llama-3.1-8B-Instruct & 36.15\% & 36.92\% \\
Gemma-2B-IT & 7.69\% & 7.69\% \\
Gemma-1.1-2B-IT & 8.46\% & 6.92\% \\
Gemma-7B-IT & 2.31\% & 4.62\% \\
Gemma-1.1-7B-IT & 0.77\% & 3.85\% \\
Mistral-7B-v0.1 & 26.15\% & 30.00\% \\
Mistral-7B-v0.2 & 26.15\% & 32.31\% \\
Mistral-7B-v0.3 & 26.15\% & 33.08\% \\
Mistral-7B-Instruct-v0.1 & 15.38\% & 20.77\% \\
Mistral-7B-Instruct-v0.2 & 31.54\% & 32.31\% \\
Mistral-7B-Instruct-v0.3 & 39.23\% & 35.38\% \\
Yi-6B & 3.85\% & 29.23\% \\
Yi-6B-Chat & 20.77\% &30.77\% \\
Zephyr-7B-Alpha & 23.08\% & 23.08\% \\
Zephyr-7B-Beta & 17.69\% & 14.62\% \\
GPT-3.5-Turbo & 37.69\% & 38.46\% \\
GPT-4 & \underline{55.38\%} & \underline{53.85\%} \\
GPT-4-Turbo & \textbf{56.92\%} & \textbf{56.15\%} \\
GPT-4o & 42.31\% & 45.38\% \\
GPT-4o-mini & 47.69\% & 46.15\% \\
\bottomrule
\end{tabular}
\caption{Comparison of the accuracy of multiple large language models on the EconLogicQA dataset under 1-shot and 5-shot learning scenarios.}
\label{tab:results}
\end{table}

These experiments collectively demonstrate the varying degrees of proficiency in applying LLMs to economic sequential reasoning, reflecting the current landscape of LLMs capabilities in this domain. There is still a clear gap in the ability of current LLMs to accurately handle many economic scenarios and correctly sequence events, especially open LLMs. This limitation points to significant challenges that persist in the field, emphasizing the need for targeted improvements and innovations in future research. Addressing these shortcomings could lead to more robust models that are better equipped to navigate the complexity of economic reasoning.

\subsection{Sensitivity Analyses}

This section examines how changes in event order, the number of examples (shots), and the use of Chain-of-Thought (CoT) reasoning \citep{wei2022chain} influence the performance of GPT-4 and GPT-4-Turbo, revealing their varying sensitivities to these factors. As shown in Table \ref{tab:reorder}, GPT-4 maintains steady performance even when event order is altered, especially in the 5-shot setting, while GPT-4-Turbo experiences a decline in accuracy when the order is permuted in the 1-shot scenario. This highlights GPT-4's robustness in maintaining accuracy under varying conditions, particularly when provided with more context.

\begin{table}[h!]
\centering
\begin{tabular}{lcrr}
\toprule
\textbf{Model} & \textbf{Permuted} & \textbf{1-Shot} & \textbf{5-Shot} \\
\midrule
GPT-4 & \xmark & 55.38\% & 53.85\% \\
GPT-4 & \cmark & 52.31\% & 53.08\% \\
GPT-4-Turbo & \xmark & 56.92\% & 56.15\% \\
GPT-4-Turbo & \cmark & 47.69\% & 50.77\% \\
\bottomrule
\end{tabular}
\caption{Accuracy of GPT-4 and GPT-4-Turbo on the EconLogicQA dataset with and without permuted order under 1-shot and 5-shot learning scenarios.}
\label{tab:reorder}
\end{table}

Table \ref{tab:few-shot} and Table \ref{tab:cot} further explore the impact of example quantity and CoT reasoning. Performance peaks at the 1-shot and 5-shot settings, with accuracy dropping as more than five examples are introduced, suggesting that excessive context may introduce confusion rather than clarity. CoT reasoning enhances performance in the 5-shot setting by promoting better coherence and logical flow, but in the 1-shot scenario, it has minimal impact, indicating that the model’s pre-trained knowledge is often sufficient for simpler tasks without needing explicit reasoning chains.

\begin{table}[h!]
\centering
\begin{tabular}{lcr}
\toprule
\textbf{Model} & \textbf{\# of Shots} & \textbf{Accuracy} \\
\midrule
GPT-4-Turbo & 0 & 48.46\% \\
GPT-4-Turbo & 1 & 56.92\% \\
GPT-4-Turbo & 2 & 51.54\% \\
GPT-4-Turbo & 3 & 53.08\% \\
GPT-4-Turbo & 4 & 51.54\% \\
GPT-4-Turbo & 5 & 56.15\% \\
GPT-4-Turbo & 8 & 50.00\% \\
GPT-4-Turbo & 10 & 50.00\% \\
\bottomrule
\end{tabular}
\caption{Accuracy of GPT-4-Turbo on the EconLogicQA dataset with different numbers of shots.}
\label{tab:few-shot}
\end{table}

\begin{table}[h!]
\centering
\begin{tabular}{lcrrr}
\toprule
\textbf{Model} & \textbf{CoT} & \textbf{1-Shot} & \textbf{5-Shot} \\
\midrule
GPT-4 & \xmark & 55.38\% & 53.85\% \\
GPT-4 & \cmark & 55.38\% & 54.62\%  \\
GPT-4-Turbo & \xmark & 56.92\% & 56.15\% \\
GPT-4-Turbo & \cmark & 55.38\% & 59.23\% \\
\bottomrule
\end{tabular}
\caption{Accuracy of GPT-4 and GPT-4-Turbo on the EconLogicQA dataset with and without Chain-of-Thought (CoT) Reasoning under 1-shot and 5-shot learning scenarios.}
\label{tab:cot}
\end{table}

\section{Conclusion}

This study introduces EconLogicQA, a novel benchmark specifically designed to assess the logical reasoning capabilities of large language models (LLMs) in the domains of economics, business, and supply chain management. The benchmark challenges LLMs with complex realistic economic scenarios. Utilizing GPT-4, high-quality, well-crafted multiple-choice questions are generated from business articles and refined through meticulous human review. A comprehensive evaluation of both open and proprietary LLMs is conducted, providing deep insights into their capabilities and limitations within this specialized context. The error analysis in Appendix \ref{sec:error} reveals that the models often struggle with sequencing strategic decisions, interpreting policy changes, and distinguishing between stages of implementation and development, indicating a need for improvement in handling complex economic reasoning.

Future improvements to LLMs in economic reasoning could include refining prompt engineering to enhance accuracy in complex scenarios and applying fine-tuning with the EconLogicQA dataset for optimized responses. Additionally, developing specialized LLMs designed for economics, business, and supply chain management offers a promising opportunity.

\section*{Limitations}

\textbf{Scope of Data.} The effectiveness of the EconLogicQA benchmark is currently validated using a specific dataset of economic news articles. This reliance on a single data source limits the generalizability of our findings to other datasets in the domain, which may have distinct characteristics and diverse compositions that could influence the performance of LLMs.

\textbf{Temporal Coverage.} The dataset spans articles from 2011 to 2022, potentially missing recent economic developments and trends. This temporal limitation could affect the benchmark's relevance and the models' performance in current economic contexts.

\section*{Ethical Considerations}

Our research emphasizes transparency in methodology, reporting, and data utilization. We adhered to the principles of responsible AI research throughout the study. The data employed in this research is sourced from public domains, ensuring no private user data was involved. We incorporated a stringent human review process to maintain dataset accuracy and integrity, excluding any sensitive or inappropriate content. These measures reflect our commitment to ethical standards and research integrity.

\bibliography{custom}

\appendix
\newpage

\section{Prompt and Response Example}
\label{sec:prompt}

Figure \ref{fig:prompt} illustrates an example of the prompt used to generate the sorting question and the corresponding response from GPT-4.

\begin{figure*}[!h]
  \centering
  \begin{tcolorbox}
    \textbf{Prompt:}
    
    Generate a concise multi-choice question without referencing the original news article, starting with a brief scenario description. Include four events from this scenario that need sequencing based on their logical connection, not just the order of appearance in the news. These events should be relevant to typical business or economic situations, requiring knowledge of business practices and economics for correct sequencing. Ensure a single correct sequence, free from ambiguity. The question should be crafted for independent completion, suitable for student evaluation. Create original, concise content for both the question and choices, avoiding plagiarism and limiting excessive details or numerical data. Format the output with the scenario description, followed by choices labeled A, B, C, D, and finish with the correct sequence and a brief explanation, ensuring the sequence logic is deducible solely from the question and choices presented.\\

    Here is the output format:

    \verb|```|\\
    Question: ...\\
    Choices:\\
    A. ...\\
    B. ...\\
    C. ...\\
    D. ...\\
    Correct Order: ...\\
    Explanation: ...\\
    \verb|```|\\

    Here is the business news article:
    
    \verb|```| \\
    \{news\_article\} \\
    \verb|```|

    \textbf{GPT-4 Response:}

    Question: Arrange the following events related to the development and launch of Apple's Siri voice assistant in their correct chronological order.\\
    
    Choices:\\
    A. Apple acquired Siri in April 2010.\\
    B. Siri was spun out as a standalone business in 2008.\\
    C. Siri was integrated into the iPhone 4S, which was announced to be sold in stores on October 14.\\
    D. Work on Siri began in the summer of 2007 at SRI International.\\
    
    Correct Order: D, B, A, C\\
    
    Explanation: The development of Siri started in the summer of 2007 at SRI International. The project was then spun out as a standalone business in 2008. Apple acquired Siri in April 2010. Finally, Siri was integrated into the iPhone 4S, which was announced to be sold in stores on October 14.

  \end{tcolorbox}
  \caption{The prompt structure for generating sorting questions in EconLogicQA. It involves filling the contents of news articles into prompt templates. We also give an example of GPT-4 response to a specific prompt constructed from this information.}
  \label{fig:prompt}
\end{figure*}

\section{Error Analyses}
\label{sec:error}

To systematically understand the errors made by the LLMs, we implement a series of automated analyses using a set of specially designed prompts. These prompts guide GPT-4o in identifying and categorizing errors based on their types and underlying causes. The instruction prompts are divided into four parts: Overview, Provided Materials, Steps to Follow, and Error Types and Criteria. The Overview provides a general introduction, the Provided Materials section lists the necessary resources, and the Steps to Follow outlines the process for using the prompts. The Error Types and Criteria part details the specific error categories, which include cause and effect misinterpretation, strategic and decision-making confusion, implementation and development confusion, policy and regulatory misunderstanding, impact and outcome misinterpretation, and growth and expansion misinterpretation. Each error type is matched with specific indicators to ensure precise classification, enhancing the clarity and accuracy of the analysis. For more specific details, please refer to Appendix \ref{sec:review}.

\begin{figure}
    \centering
    \includegraphics[width=.8\linewidth]{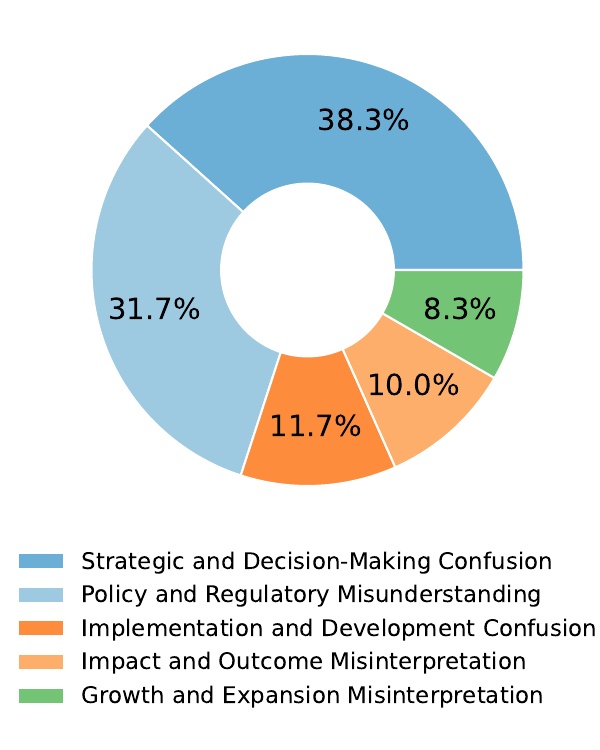}
    \caption{Error types of GPT-4 responses on the testing set.}
    \label{fig:pie}
\end{figure}

According to Figure \ref{fig:pie}, the error analysis reveals that the most common error type is strategic and decision-making confusion, which accounts for 38.3\% of the total errors. This indicates that the model often struggles with understanding the correct sequence of strategic decisions and their logical order. The second most frequent error type is policy and regulatory misunderstanding, making up 31.7\% of errors, highlighting the model's difficulty in correctly interpreting the sequence and impact of policy changes. Implementation and development confusion constitutes 11.7\% of the errors, reflecting challenges in distinguishing between various stages of project execution. Impact and outcome misinterpretation account for 10\% of the errors, showing some issues in linking actions with their results. Lastly, growth and expansion misinterpretation, comprising 8.3\% of the errors, indicates some misunderstanding in the sequence of growth-related activities. These insights suggest that while the model performs reasonably well in some areas, significant improvements are needed in understanding strategic decisions and policy sequences.

\section{Instruction Sheet for Exam Review}
\label{sec:review}

\textbf{Overview}

This instruction sheet provides guidance on how to review and classify responses in an economics question set. The exam requires sorting choices correctly, and the provided responses contain various errors. The goal is to classify each response into its respective error type using the detailed criteria provided.

\textbf{Provided Materials}

\begin{enumerate}
    \item Exam Questions and Choices: Each question includes multiple choices that need to be sorted in the correct sequence.
    \item Correct Answers: The correct order for each question is provided.
    \item Responses: The responses that need to be reviewed and classified according to the error types.
\end{enumerate}

\textbf{Steps to Follow}

\begin{enumerate}
    \item Read the Response: Carefully read the response to the question.
    \item Identify Misunderstandings: Look for signs of misunderstanding or confusion in the response.
    \item Match Indicators: Compare the mistakes to the indicators listed in each error type.
    \item Classify the Error: Assign the mistake to the appropriate error type based on the matching indicators.
    \item Provide Feedback: Offer specific feedback based on the identified error type to help understand and correct the mistake.
\end{enumerate}

\textbf{Error Types and Criteria}
\begin{enumerate}
    \item Cause and Effect Misinterpretation
    \begin{itemize}
        \item Description: Errors where the relationships between causes and their effects, including immediate versus delayed reactions, are misunderstood.
        \item Indicators:
        \begin{itemize}
            \item An effect is placed before its cause.
            \item Confusion between actions and their immediate or delayed outcomes.
        \end{itemize}
        \item Examples:
        \begin{itemize}
            \item Misinterpreting which actions lead to specific results.
            \item Placing the result of an action before the action itself.
        \end{itemize}
    \end{itemize}

    \item Strategic and Decision-Making Confusion
    \begin{itemize}
        \item Description: Errors involving misunderstanding strategic decisions and the correct sequence of these decisions.
        \item Indicators:
        \begin{itemize}
            \item Confusion about the order of strategic steps.
            \item Misordering company actions or milestone events.
        \end{itemize}
        \item Examples:
        \begin{itemize}
            \item Placing a company's strategic milestone after subsequent actions.
            \item Misunderstanding the sequence of decision-making processes.
        \end{itemize}
    \end{itemize}

    \item Implementation and Development Confusion
    \begin{itemize}
    \item Description: Errors related to confusion between different stages of implementation, development, and testing processes.
    \item Indicators:
    \begin{itemize}
        \item Mixing up the order of development stages.
        \item Placing testing phases incorrectly.
    \end{itemize}
    \item Examples:
    \begin{itemize}
        \item Placing the implementation stage before the development stage.
        \item Misordering feature rollouts and testing phases.
    \end{itemize}
    \end{itemize}

    \item Policy and Regulatory Misunderstanding
    \begin{itemize}
        \item Description: Errors involving misunderstandings of policy changes, regulatory actions, and economic indicators.
        \item Indicators:
        \begin{itemize}
            \item Misinterpreting the sequence of policy changes.
            \item Confusion about the order of regulatory actions.
        \end{itemize}
        \item Examples:
        \begin{itemize}
            \item Misplacing the impact of a policy change.
            \item Misunderstanding the sequence of regulatory steps.
        \end{itemize}
    \end{itemize}

    \item Impact and Outcome Misinterpretation
    \begin{itemize}
        \item Description: Errors related to misinterpreting the impact or outcomes of actions and decisions.
        \item Indicators:
        \begin{itemize}
            \item Confusion between actions and their impacts.
            \item Misplacing outcomes in the sequence of events.
        \end{itemize}
        \item Examples:
        \begin{itemize}
            \item Placing an outcome before the action that caused it.
            \item Misunderstanding the effect of a specific action or decision.
        \end{itemize}
    \end{itemize}

    \item Growth and Expansion Misinterpretation
    \begin{itemize}
        \item Description: Errors related to misunderstanding growth metrics and expansion steps.
        \item Indicators:
        \begin{itemize}
            \item Misordering steps related to company growth.
            \item Confusion about the sequence of expansion activities.
        \end{itemize}
        \item Examples:
        \begin{itemize}
            \item Placing growth milestones out of order.
            \item Misunderstanding the sequence of steps for company expansion.
        \end{itemize}
    \end{itemize}

\end{enumerate}

\textbf{Example Application}

Question: The Federal Reserve is considering the creation of a digital currency. What steps did they take in sequence?

Choices:

A: The Federal Reserve collects public feedback on the digital currency.

B: The Federal Reserve releases a discussion paper on the digital currency.

C: The Federal Reserve receives clear support from the public.

D: The Federal Reserve addresses major policy questions.

Correct Answer: B, A, C, D

Response: B, A, D, C

Error Type: Policy and Regulatory Misunderstanding

Indicators Matched: Misinterprets the sequence of policy changes (placing policy questions before public support).

Feedback: The response indicates a misunderstanding of the sequence in which the Federal Reserve addressed policy questions and gathered public support. It's important to understand that the Federal Reserve first gathers public feedback before addressing policy questions.

\end{document}